%% file: acl2021.tex
\documentclass[11pt,a4paper]{article}
\usepackage[hyperref]{acl2021}
\usepackage{times}
\usepackage{latexsym}

\usepackage{graphicx}
\usepackage{booktabs}
\usepackage{amsmath}
\usepackage{subcaption}
\usepackage{stfloats}

\usepackage{microtype}

\aclfinalcopy 
\def\aclpaperid{15} 

\title{Evaluating Compositional Generalisation in VLMs and Diffusion Models}

\author{
  Beth Pearson \\
  University of Bristol \\
  \texttt{\small{beth.pearson@bristol.ac.uk}}
  \And
  Bilal Boulbarss \\
  University of Amsterdam \\
  \texttt{\small{bilal.boulbarss@student.uva.nl}}
  \AND
  Michael Wray \\
  University of Bristol \\
  \texttt{\small{michael.wray@bristol.ac.uk}}
  \And
  Martha Lewis \\
  University of Amsterdam \\
  \texttt{\small{m.a.f.lewis@uva.nl}}
}

\date{}

\begin{document}
\maketitle

\input{content.tex}

\end{document}

%% file: content.tex
\begin{abstract}
A fundamental aspect of the semantics of natural language is that novel meanings can be formed from the composition of previously known parts.
Vision-language models (VLMs) have made significant progress in recent years, however, there is evidence that they are unable to perform this kind of composition. For example, given an image of a red cube and a blue cylinder, a VLM such as CLIP is likely to incorrectly label the image as a red cylinder or a blue cube, indicating it represents the image as a `bag-of-words' and fails to capture compositional semantics. Diffusion models have recently gained significant attention for their impressive generative abilities, and zero-shot classifiers based on diffusion models have been shown to perform competitively with CLIP in certain compositional tasks. In this work we explore whether the generative Diffusion Classifier has improved compositional generalisation abilities compared to discriminative models. We assess three models---Diffusion Classifier, CLIP, and ViLT---on their ability to bind objects with attributes and relations in both zero-shot learning (ZSL) and generalised zero-shot learning (GZSL) settings. Our results show that the Diffusion Classifier and ViLT perform well at concept binding tasks, but that all models struggle significantly with the relational GZSL task, underscoring the broader challenges VLMs face with relational reasoning. Analysis of CLIP embeddings suggests that the difficulty may stem from overly similar representations of relational concepts such as left and right. Code and dataset are available at: \href{https://github.com/otmive/diffusion_classifier_clip}{github.com/otmive/diffusion\_classifier\_clip}

\end{abstract}

\section{Introduction}
\label{sec:intro}
\begin{figure}[htbp]
    \centering
    \includegraphics[width=0.9\columnwidth]{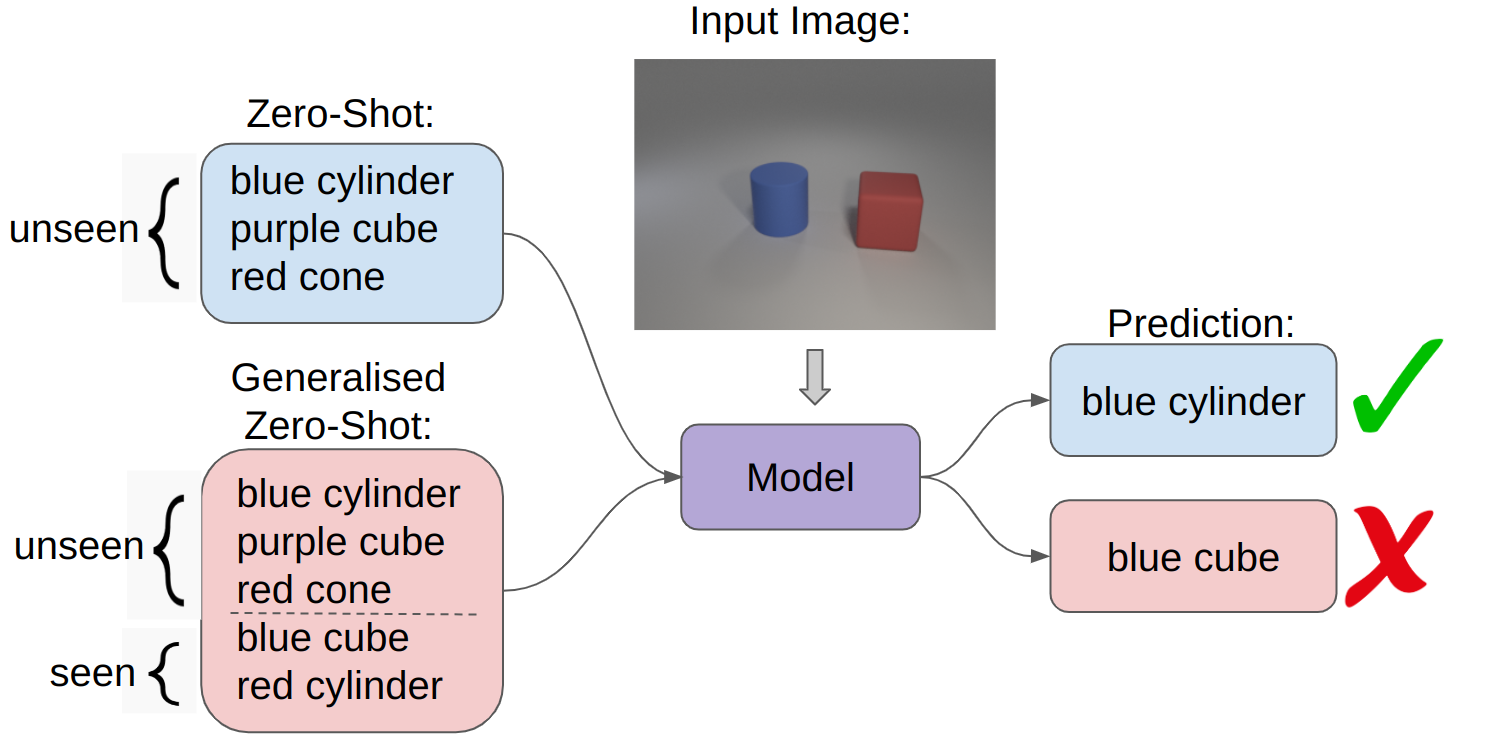}
    \caption{We evaluate the compositional generalisation of Vision-Language Models (VLMs) by assessing their ability to bind colours to objects and relations to objects in both zero-shot and generalised zero-shot settings across single-object, two-object, and relational scenarios}
    \label{fig:diagram}
\end{figure}
Compositionality is a fundamental part of how humans learn~\cite{reason:Chomsky57a, janssen1997compositionality}. It allows us to take familiar concepts and combine them in new ways to interpret novel situations, learn from limited examples, and build increasingly complex ideas. However, vision-language models (VLMs) fall short in tasks requiring compositional understanding~\cite{diwan-etal-2022-winoground, yuksekgonul2022and, lewis-etal-2024-clip}. Even with advances in attention mechanisms~\cite{vani2024sparo} and positional encoding~\cite{su2024roformer}, VLMs are unable to match the compositional reasoning skills of humans \cite{sinha2024survey, hua2024mmcomposition}. 
VLMs such as CLIP~\cite{radford2021learning} have been shown to treat captions as a bag-of-words~\cite{thrush2022winoground} and are not able to bind concepts to objects in the same way humans can. For example, given an image of a \emph{red cube} and a \emph{blue cylinder}, a VLM may misinterpret the image as containing a \emph{blue cube} or a \emph{red cylinder} (see Figure \ref{fig:diagram}). Additionally, a VLM should be able to generalise learned concepts to new unseen combinations of attributes and objects: if a model learns the colour \emph{cyan} through images of \emph{cyan cone} and the shape \emph{cube} through images of \emph{green cubes}, it should also be able to recognise images of \emph{cyan cubes} or \emph{green cones}. 

Diffusion Models have gained significant interest in recent years for their state-of-the-art performance on image generation~\cite{ramesh2022hierarchical, dhariwal2021diffusion} and editing tasks~\cite{brooks2023instructpix2pix}. Their performance as zero-shot classifiers in vision tasks is a recent topic of exploration~\cite{clark2023text,krojer2023diffusion}. On compositional benchmarks such as Winoground \cite{thrush2022winoground} or the Concept Binding Benchmark from \citet{lewis-etal-2024-clip}, their performance has been shown to be comparable to that of CLIP \cite{li2023your,clark2023text}. However, Winoground has been argued to require commonsense and world knowledge rather than purely testing for compositional abilities \cite{diwan-etal-2022-winoground}, and performance on the Concept Binding Benchmark can be at chance.

In this paper, we contribute to the understanding of the compositional abilities of diffusion model-based classifiers by comparing with Transformer-based classifiers on compositional tasks. Specifically, we explore how these two types of models are able to compose attributes and relations---tasks VLMs particularly struggle with. We aim to assess whether Diffusion Classifier can offer new insights or improvements in handling these challenging aspects of compositional semantics.

We consider two settings for our experiments---zero-shot learning (ZSL) and generalised zero-shot learning (GZSL). In ZSL, the aim is to recognise only unseen classes whereas GZSL aims to train models that are able to discriminate between both seen and unseen classes during test time~\cite{pourpanah2022review, xian2017zero}. The GZSL setting is particularly important for real world scenarios as there may only be labelled data for a small number of classes and capturing every possible class in the training set is often impossible. Therefore, it is important for models to be able to generalise to unseen classes in the presence of labels that have previously been seen.

To probe these abilities, we extend the Concept Binding Benchmark from \citet{lewis-etal-2024-clip}, which evaluates model performance on attribute-object binding and relational composition. We evaluate the performance of Diffusion Classifier---a classifier built from Stable Diffusion~\cite{rombach2022high}---comparing it with CLIP and ViLT~\cite{kim2021vilt}. Despite the dataset being lightweight, it still proves a challenge for the models, particularly in the important GZSL setting.

We outline the dataset design, discuss the experimental setup, and provide an analysis of the model's performances. We then analyse CLIP, ViLT, and Stable Diffusion's understanding of relational concepts. We find that ViLT has the strongest performance in both two-object tasks and Diffusion Classifier outperforms CLIP on out-of-distribution examples demonstrating less over-fitting to the training data. However, all models have difficulty in the GZSL relational setting and cannot differentiate between left and right relations. 

The main contributions of this work are threefold:
\textbf{(1)} We compare Diffusion Classifier, CLIP, and ViLT on compositional tasks. Diffusion Classifier generalises best in single-object settings, however, ViLT has by far the best two-object performance. All models struggle to reliably compose relations with objects.
\textbf{(2)} To provide a more robust evaluation of compositional generalisation, we present our extension of the Concept Binding Benchmark from \citet{lewis-etal-2024-clip}. This extended benchmark consists of three datasets to test Vision-Language Models in both zero-shot (ZSL) and generalised zero-shot (GZSL) scenarios.
\textbf{(3)} We analyse the effects of fine-tuning on compositional semantic understanding, showing that models fail to form disentangled representations for spatial relations.

\section{Related Work}
\label{sec:relatedwork}
\paragraph{Benchmarking Compositionality in VLMs}
There is a growing interest in the ability of VLMs to reason compositionally, with several benchmarks being proposed in recent years~\cite{yuksekgonul2022and, ma2023crepe, hsieh2024sugarcrepe, dumpala2024sugarcrepe++, ray2024cola, zhao2022vl, huang2024conme, thrush2022winoground, hua2024finematch}. Compositional generalisation is an important ability for VLMs to have because it encourages the interpretability and data efficiency of models~\cite{bommasani2021opportunities}. However, it has been argued \cite{lewis-etal-2024-clip,hsieh2024sugarcrepe} that various compositionality benchmarks are `hackable', showing that in some cases it is possible to solve the benchmark simply by comparing prompts \cite{wu23a} and ignoring the image. SugarCrepe \cite{hsieh2024sugarcrepe} is designed to deal with this problem, but is still prone to the issue that the correct caption is statistically more likely in the training corpus. Unlike benchmarks that use complex real-world images, we use simple, synthetic images to ensure no spurious correlations and to directly test compositional understanding. We argue that VLMs should be able to handle these simpler reasoning tasks before advancing to more complex, real-world images.

\paragraph{Improving Compositionality in VLMs}
Methods have been proposed to improve the compositional abilities of VLMs~\cite{cascante2023going, doveh2023teaching}. Several works use hard negative sampling to fine-tune CLIP on batches of similar images e.g. ``a black cat sitting on a desk'' and ``a black desk sitting on a cat'' which force the model to learn more detailed representations of the data~\cite{yuksekgonul2022and, shou2024enhancing, sahin2024enhancing}.  Other methods include different representations for objects within images such as trees or graphs~\cite{singh2023coarse, yellinek20233vl} and adaptations to the contrastive loss function of CLIP to include more compositional supervision~\cite{pandey2022cross, zhang2024contrasting}. Despite advancements, VLMs still struggle with compositional reasoning~\cite{hsieh2024sugarcrepe, dumpala2024sugarcrepe++}. Our benchmark aims to investigate why VLMs struggle with compositional tasks by testing in GZSL settings using in-distribution and OOD images to identify potential biases.

\paragraph{Diffusion Model Classifiers}
Recently, methods have been proposed to leverage diffusion models as zero-shot classifiers~\cite{chen2023robust, li2023your, krojer2023diffusion, clark2023text}.
\citet{li2023your} propose Diffusion Classifier, a model built from Stable Diffusion, which achieves a higher accuracy than CLIP on tasks requiring compositional reasoning such as concept binding.
\citet{krojer2023diffusion} use a similar method for using Stable Diffusion~\cite{rombach2022high} as a classifier but include a normalising value based on the noise prediction error calculated with no text guidance. \citet{he2023discriminative} use the attention scores between the image and text representations of Stable Diffusion to adapt it for image-text matching tasks. 
\citet{clark2023text} also propose a zero-shot classifier created from Google's Imagen, which shows some ability to bind attributes such as shape, size and colour where CLIP fails to do so. For our experiments we use the Diffusion Classifier from \citet{li2023your} as Stable Diffusion is open source with easily accessible fine-tuning methods. 

\section{Experiments}
We base the design of our benchmark on the experiments from~\citet{lewis-etal-2024-clip} where three datasets were created for exploring composition of attributes and relations with objects. While this setup reveals that models often struggle even with simple object compositions, our aim is to extend this evaluation to include both Zero-Shot Learning (ZSL) and Generalised Zero-Shot Learning (GZSL) settings. To enable this, we adapt and expand the original benchmark to support systematic and rigorous testing in both settings.

\begin{figure}
    \centering
    \includegraphics[width=0.9\columnwidth]{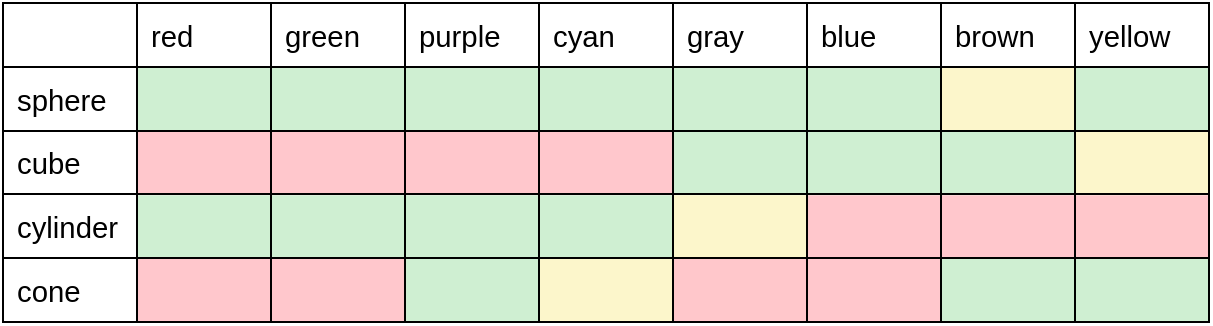}
    \caption{Single and Two-Object dataset design. Class labels belonging to each dataset split: train and in-distribution are highlighted in green, OOD validation in yellow, and OOD test in red. }
    \label{fig:dataset_split}
\end{figure}

\begin{figure}[htbp]
    \centering
    \includegraphics[width=0.9\columnwidth]{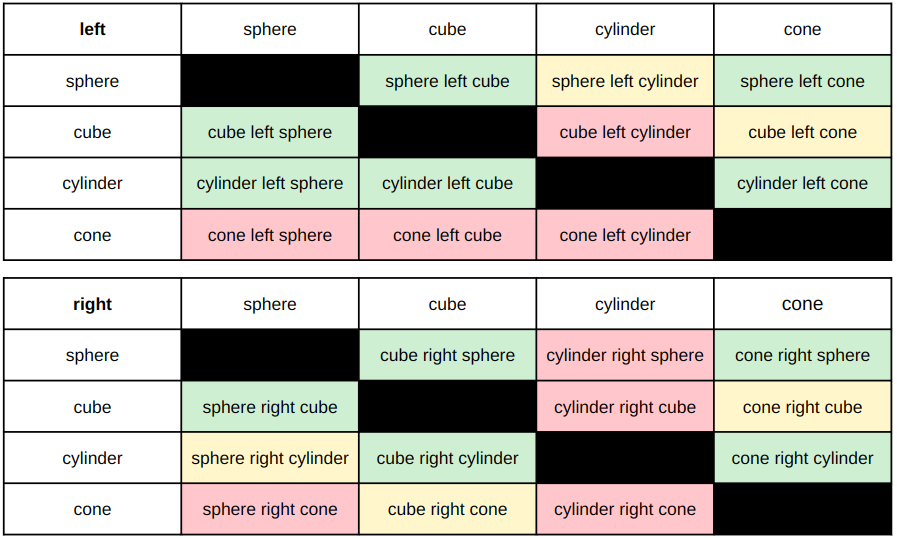}
    \caption{Relational dataset design. Class labels belonging to each dataset split: train and in-distribution are highlighted in green, OOD validation in yellow, and OOD test in red.}
    \label{fig:rel_dataset_split}
\end{figure}

The images are generated using the generation script for the CLEVR dataset~\cite{johnson2017clevr}---using a Blender script~\cite{blender} to render 3D shapes.
The original code included only three shapes \textit{cubes}, \textit{cylinders}, and \textit{spheres} which we extend with an additional shape, \textit{cones}, to increase the diversity across the dataset splits. For the single and two-object datasets, we consider the following colours: \textit{blue}, \textit{brown}, \textit{cyan}, \textit{gray}, \textit{green}, \textit{purple}, \textit{red}, and \textit{yellow}.
We define the label sets for the single and two-object datasets as follows:

Let \( C \) be the set of colours and \( S \) the set of shapes. For object classification, each object is identified by its colour--shape pair, and the label set is defined as:
\[
\mathcal{Y} = \{(c, s) \mid c \in C,\, s \in S\}.
\]
Each element of \(\mathcal{Y}\) represents a unique object (e.g., \textit{red square}, \textit{blue circle}). In the two-object dataset, labels consist of two such tuples, e.g., \(((c_1, s_1), (c_2, s_2))\).
For the relational dataset, we define a set of spatial relations \( R = \{\textit{left},\, \textit{right}\} \). We exclude the relations \textit{front} and \textit{behind} which were included in Lewis et al.~\cite{lewis-etal-2024-clip} as we found these to be too ambiguous---distinguishing which shape is further forward is often difficult even for humans. 
The relational label set is then defined as:
\[
\mathcal{Y}_{\text{rel}} = \{(s_i, r, s_j) \mid s_i, s_j \in S,\, s_i \neq s_j,\, r \in R\},
\]
where each triple describes a relation between two distinct shapes---for example, \((\textit{circle}, \textit{left}, \textit{square})\).
All datasets are partitioned into five subsets: training (\(\mathcal{Y}^{\text{train}}\)), in-distribution validation/test (\(\mathcal{Y}^{\text{IDval}}, \mathcal{Y}^{\text{IDtest}}\)), and out-of-distribution validation/test (\(\mathcal{Y}^{\text{OODval}}, \mathcal{Y}^{\text{OODtest}}\)). In-distribution subsets use the same label space as the training set:
\[
\mathcal{Y}^{\text{train}} = \mathcal{Y}^{\text{IDval}} = \mathcal{Y}^{\text{IDtest}},
\]
while OOD splits are defined such that:
\[
\begin{aligned}
\mathcal{Y}^{\text{train}} \cap \mathcal{Y}^{\text{OODval}} &= \emptyset, \quad
\mathcal{Y}^{\text{train}} \cap \mathcal{Y}^{\text{OODtest}} = \emptyset, \\
\mathcal{Y}^{\text{OODval}} \cap \mathcal{Y}^{\text{OODtest}} &= \emptyset.
\end{aligned}
\]
This setup enables evaluation both within the training distribution and on novel combinations, to assess generalisation.
We give the structure of our single and two-object datasets within  Figure~\ref{fig:dataset_split}. The label \textit{red cube} is in the test set, meaning that it is not seen during training, but \textit{red} (e.g. in \textit{red sphere}) and \textit{cube} (e.g. in \textit{gray cube}) have both been seen during training in other combinations. The structure of the relational dataset is given in Figure~\ref{fig:rel_dataset_split}.
For both ZSL and GZSL tasks, models are trained on images and labels from the training split of the data. In the ZSL setting, at test time, models must pick the correct label for an image from a set $\mathcal{S}$ of unseen labels, i.e. $\mathcal{S}\subseteq \mathcal{Y}^{\text{OODtest}}$. In the GZSL task, at test time, models must pick the correct label for an image from a set of both seen and unseen labels, i.e. $\mathcal{S}\subseteq \mathcal{Y}$ or $\mathcal{S}\subseteq \mathcal{Y_{\text{rel}}}$. This setup evaluates the ability of models fine-tuned on the train split to generalise colours or relations learned during training to new unseen shape combinations.
Because of this, the single and two-object train split contains at least one class containing each shape and each colour. Similarly, the relational train split contains at least one of each shape. 

We only use positive examples when fine-tuning CLIP rather than both positive and negative examples to keep consistent with the DreamBooth fine-tuning method for Stable Diffusion which only accepts positive training examples. In addition, to further align with DreamBooth, we fine-tune CLIP with a small number of samples from each class (20-40 per class).

\begin{table}
\centering
\resizebox{\columnwidth}{!}{%
    \begin{tabular}{c c c c c c}
    \hline
         &\textbf{Train} & \textbf{ID Val} & \textbf{ID Test} & \textbf{OOD Val} & \textbf{OOD Test}  \\
         \hline
         Single-Object & 1360 & 340 & 340 & 400 & 1100 \\
         Two-Object & 7440 & 1860 & 1860 & 600 & 3700 \\
         Relational & 440 & 110 & 110 & 250 & 400 \\\hline
         
    \end{tabular}
    }
    \caption{Our extended benchmark statistics for the three datasets showcasing the number of images within each of the splits.}
\label{tab:dataset_statistics}
\end{table}

\subsection{Single-Object}
The single-object task tests the model's ability to recognise attribute-object pairs and is used as a baseline for analysing which combinations the models can recognise before experimenting in a two-object setting.
Examples from the single-object dataset are shown in Figure~\ref{dataset_examples} a) and b). In the single-object setting, we evaluate only on the GZSL task, and require models to select the correct label for the image from all possible label combinations, i.e. from the whole of $\mathcal{Y}$.
Following convention, the class labels are given in the form of a prompt ``a photo of a \textless class\textgreater''.  

\subsection{Two-Object}
The two-object dataset contains images of exactly two-objects which differ in \emph{both shape and colour}. For example, the dataset contains images of a \textit{blue cube} and
a \textit{red sphere} but not of a \textit{blue cube} and a \textit{blue sphere}. We follow~\citet{lewis-etal-2024-clip} and present the model with labels for individual objects whereby the true label correctly describes one of the objects in the image and the others are incorrect. In comparison to giving the model a full description of the image (e.g. \textit{green cone and purple cylinder}),  this is a challenging setup which minimises the use of shortcuts by the model, for example if the model can recognise green cones correctly but not purple cylinders. As an example, the images in Figure~\ref{dataset_examples} c) may have the true label \emph{green cylinder} and hard negatives \emph{green cone} and \emph{purple cylinder}.

In the ZSL setting, models are given one correct label and two distractors from the same (unseen) split. For example, an image of a yellow cube may be paired with gray cylinder and brown sphere as distractors (see Figure~\ref{dataset_examples} column d)).

In the GZSL setting, models choose from five labels: the true label, two standard distractors, and two hard negatives created by swapping attributes and shapes (e.g., yellow cone, cyan cube for Figure \ref{dataset_examples} d)). This makes the task more challenging and tests whether models prefer familiar (seen) classes over novel ones.

\subsection{Relational}
\begin{figure*}[htbp]
\centering
\includegraphics[width=2\columnwidth]{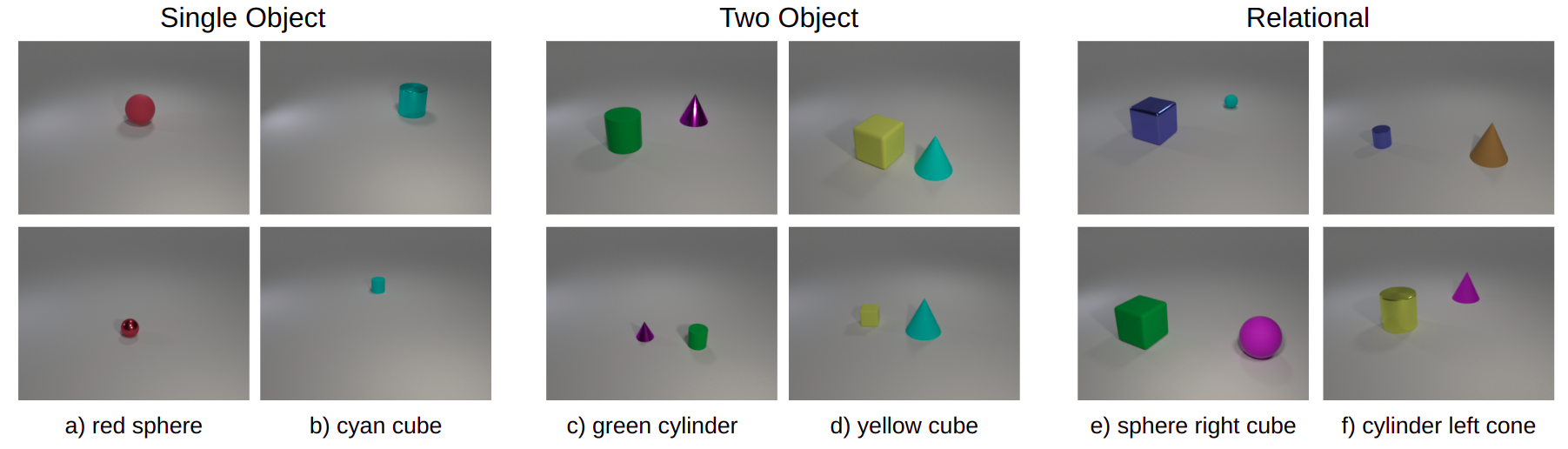}
\caption{Samples from our extended benchmark with two example classes displayed from each dataset---single, two-object, and relational. }
\label{dataset_examples}
\end{figure*}

The relational dataset tests compositions of the relations \textit{left} and \textit{right} between two-objects in an image. The two-objects are always two distinct shapes, that is, we don't consider cases such as \textit{sphere left sphere}. As with the two-object dataset, each image has two possible true labels. For instance, the images in Figure~\ref{dataset_examples} column e) would have the true labels \textit{cube left sphere} and \textit{sphere right cube}. Again, we consider a ZSL and GZSL setting. In the GZSL setting, models choose from five options: the true label, two randomly selected labels, and two hard negatives. One hard negative alters the spatial relation (e.g., cube left sphere $\to$ cube right sphere), while the other swaps object order (e.g., cube left sphere $\to$ sphere left cube). The hard negatives require the model to recognise the specific relation in the image and not just recognise which two shapes are present---a task at which a bag-of-words model would fail. 

\section{Results}
We conduct experiments comparing frozen and fine-tuned CLIP, ViLT, and Diffusion Classifier (DC) on three datasets: single-object, two-object, and relational. Experiments are carried out in a Linux environment using an RTX 2080 GPU for both training and inference. 

\subsection{Single-Object}
For each of the models we fine-tune with three different seeds and report the mean and standard deviation of each. Fine-tuning details and hyperparameters for each dataset are provided in Appendix \ref{app:fine-tune}.

 \paragraph{Results}
 \begin{table}[htbp]
  \begin{center}
  \resizebox{\columnwidth}{!}{%
    \begin{tabular}{lccccccr}
    \toprule
    \textbf{Model} & \textbf{ID Validation} & \textbf{ID Test} & \textbf{OOD Validation} & \textbf{OOD Test}\\\midrule
    Frozen CLIP & 85.29\textsuperscript{0.00} & 80.59\textsuperscript{0.00} & 67.75\textsuperscript{0.00} & 87.36\textsuperscript{0.00} \\
    CLIP-FT & 95.29\textsuperscript{3.01} & 95.59\textsuperscript{2.92} & 93.57\textsuperscript{3.81} & 91.21\textsuperscript{6.54} \\
    Frozen ViLT & 51.47\textsuperscript{0.00} & 50.0\textsuperscript{0.00} & 34.5\textsuperscript{0.00} & 44.91\textsuperscript{0.00}\\
    ViLT-FT & 95.88\textsuperscript{0.00} & 94.71\textsuperscript{0.00} & 63.5\textsuperscript{0.00} & 77.18\textsuperscript{0.00} \\
    Frozen DC & 40.80\textsuperscript{0.89} & 40.98\textsuperscript{0.37}& 58.0\textsuperscript{0.50} & 60.0\textsuperscript{1.08} \\
    DC-FT & 97.74\textsuperscript{1.6} & 97.16\textsuperscript{0.78}& 99.50\textsuperscript{0.12} & 99.47\textsuperscript{0.87} \\ 
    \bottomrule
    \end{tabular}
    }
    \caption{Accuracy of models on the single-object task.}
    \label{tab:single-zero-shot}
  \end{center}
\end{table}

We see in Table \ref{tab:single-zero-shot} that CLIP has the best accuracy of the frozen models on this task. However, after fine-tuning, DC has the best overall accuracy. Both CLIP and DC show a strong performance on ID and OOD splits indicating that in the simple single-object setting they are able to generalise to unseen colour-shape combinations. In contrast, fine-tuned ViLT showcases strong performance only on the ID splits and shows a drop in accuracy to 63.5\% and 77.18\% on the OOD splits. ViLT frequently makes errors such as predicting \emph{blue cone} for \emph{cyan cone} or \emph{gray cube} for \emph{gray cylinder}---failing to generalise from familiar components seen during training (such as the colour cyan with a sphere, or the shape cylinder with other colours like red, green, or purple). Fine-tuned CLIP and DC are able to generalise in the single-object setting but ViLT's lower OOD performance shows even in simple settings composing unseen combinations can be difficult for VLMs.

\subsection{Two-Object Zero-Shot}
We report the average accuracy with the standard deviation for all models as shown in Table~\ref{tab:two-object-zero-shot}. 
 \begin{table}[htbp]
  \begin{center}
  \resizebox{\columnwidth}{!}{%
    \begin{tabular}{lccccccr}
    \toprule
    \textbf{Model} & \textbf{ID Validation} & \textbf{ID Test} & \textbf{OOD Validation} & \textbf{OOD Test}\\\midrule
    Frozen CLIP & 83.71\textsuperscript{0.00} & 85.27\textsuperscript{0.00} & 93.0\textsuperscript{0.00} & 69.51\textsuperscript{0.00}
    \\
    CLIP-FT & 90.13\textsuperscript{0.55}& 90.39\textsuperscript{0.01}& 99.39\textsuperscript{0.75}& 80.15\textsuperscript{1.11}\\
    Frozen ViLT & 72.78\textsuperscript{0.56} & 73.80\textsuperscript{0.73} & 70.0\textsuperscript{0.00} & 66.82\textsuperscript{0.32}\\
    ViLT-FT & 99.78\textsuperscript{0.00} & 99.89\textsuperscript{0.08} & 99.5\textsuperscript{0.00}& 99.26\textsuperscript{0.18} \\
    Frozen DC & 61.18\textsuperscript{0.00}& 64.53\textsuperscript{0.00}& 91.83\textsuperscript{0.00} & 58.3\textsuperscript{0.00}\\
    DC-FT & 82.59\textsuperscript{3.34} & 83.21\textsuperscript{3.59}& 93.89\textsuperscript{2.49} & 72.80\textsuperscript{2.06}\\ 
    \bottomrule
    \end{tabular}
    }
    \caption{Accuracy of models on the ZSL two-object task.}
    \label{tab:two-object-zero-shot}
  \end{center}
\end{table}

Frozen CLIP has the highest accuracy of the frozen models. ViLT-FT has the highest accuracy achieving over 99\% on all dataset splits. This is particularly surprising given its lower performance in the single-object task. ViLT may benefit from the reduced label space in the two-object ZSL experiment compared to having the full range of prompts in the single-object setting. CLIP-FT and DC-FT both show a decrease in performance on OOD test but not on OOD val. We believe the high OOD val accuracies are due to the reduced size of the OOD val split meaning there are only 4 very distinct prompts to choose from. The drop in performance of all models on the OOD test split further highlights that VLMs lack robust compositional understanding, even for the simpler zero-shot case. Current pre-training strategies rarely require models to explicitly learn compositional knowledge, suggesting that adjustments to pre-training may be necessary.

\subsection{Two-Object Generalised Zero-Shot}
We report the accuracies and standard deviations for the two-object GZSL experiment in Table~\ref{tab:two-object-gzs}.

 \begin{table}[htbp]
  \begin{center}
  \resizebox{\columnwidth}{!}{%
    \begin{tabular}{lccccccr}
    \toprule
    \textbf{Model} & \textbf{ID Validation} & \textbf{ID Test} & \textbf{OOD Validation} & \textbf{OOD Test}\\\midrule
    Frozen CLIP & 23.33\textsuperscript{0.00}& 21.56\textsuperscript{0.00} & 35.33\textsuperscript{0.00}&34.27\textsuperscript{0.00}
    \\
    CLIP-FT & 78.82\textsuperscript{3.05} & 75.43\textsuperscript{0.86} & 55.50\textsuperscript{5.92}& 23.38\textsuperscript{5.28}\\
    Frozen ViLT  & 31.56\textsuperscript{0.12} & 32.71\textsuperscript{0.29} & 47.83\textsuperscript{0.00} & 29.1\textsuperscript{0.21} \\
    ViLT-FT & 99.71\textsuperscript{0.07} & 99.86\textsuperscript{0.03} & 91.67\textsuperscript{0.00} & 83.46\textsuperscript{0.06}\\
    Frozen DC & 33.58\textsuperscript{0.00}& 34.64\textsuperscript{0.00} & 38.46\textsuperscript{0.00} & 39.32\textsuperscript{0.00}\\
    DC-FT & 53.06\textsuperscript{3.20} & 51.86\textsuperscript{3.41} & 57.06\textsuperscript{5.03}& 72.97\textsuperscript{2.05}\\ 
    \bottomrule
    \end{tabular}
    }
    \caption{Accuracy of models on the GZSL two-object task. }
    \label{tab:two-object-gzs}
  \end{center}
\end{table}

Again ViLT-FT has the strongest performance for all dataset splits significantly outperforming other models. This suggests it is less biased towards seen labels as evidenced by the relatively stable performance across ZSL and GZSL. However, it does still exhibit a small drop in performance on the OOD splits indicating some limitations in generalising. CLIP-FT experiences a substantial drop in performance on the OOD splits especially OOD test, showing it has over-fit to the training data. DC-FT interestingly shows the reverse pattern to the other models and has the highest accuracy on OOD. While the high OOD test accuracy is particularly notable in the challenging GZSL setting, DC's lower accuracy on the ID splits (53.06\% and 51.86\%) suggests it lacks consistent attribute-object binding ability. Even ViLT-FT, the best-performing model overall, has a reduced performance on the OOD splits, highlighting limitations in the way models represent and combine attributes and objects. 

\subsection{Relational Zero-Shot}
We show the relational ZSL results in Table~\ref{tab:relational-zero-shot}.
 \begin{table}[htbp]
  \begin{center}
  \resizebox{\columnwidth}{!}{%
    \begin{tabular}{lccccccr}
    \toprule
    \textbf{Model} & \textbf{ID Validation} & \textbf{ID Test} & \textbf{OOD Validation} & \textbf{OOD Test}\\\midrule
    Frozen CLIP & 56.36\textsuperscript{0.00}& 56.60\textsuperscript{0.00}&38.40\textsuperscript{0.00} & 68.00\textsuperscript{0.00}
    \\
    CLIP-FT & 99.39\textsuperscript{0.86}& 99.31\textsuperscript{0.57} & 68.00\textsuperscript{13.91} & 94.08\textsuperscript{3.86}\\
    Frozen ViLT & 74.55\textsuperscript{1.48} & 68.52\textsuperscript{0.87} & 42.40\textsuperscript{0.00} & 64.67\textsuperscript{0.31}\\
    ViLT-FT & 78.18\textsuperscript{2.57} & 76.04\textsuperscript{1.98} & 70.53\textsuperscript{0.19} & 65.0\textsuperscript{0.35}\\
    Frozen DC & 68.18\textsuperscript{0.00} & 69.44\textsuperscript{0.00}& 30.70\textsuperscript{0.00}& 65.25\textsuperscript{0.00}\\
    DC-FT & 89.09\textsuperscript{4.64}& 92.94\textsuperscript{1.18}& 51.86\textsuperscript{2.31}& 87.18\textsuperscript{9.18}\\ 
    \bottomrule
    \end{tabular}
    }
    \caption{Accuracy of models on the ZSL relational task.}
    \label{tab:relational-zero-shot}
  \end{center}
\end{table}

All models except ViLT-FT have a lower accuracy on OOD validation than OOD test. This could be due to the smaller size of the validation split, which limits prompt diversity making the distractor labels more likely to share shapes with the shapes in the true label. Both DC and CLIP only show slight drops in performance between OOD test and the ID splits demonstrating the capacity to recognise unseen object-relation combinations in ZSL settings. ViLT, while having overall lower accuracies, shows less variation across dataset splits, showing some capacity to generalise. All models show a substantial drop in performance in the relational ZSL compared with the two-object ZSL showing that systematically combining objects with relations is harder for these models than combining colour-object pairs. The difficulty the models have with relational information suggests they are focusing on recognising objects in the image rather than compositions between objects. While VLMs can often rely on these shortcuts and still achieve a strong performance, tasks that require relational reasoning reveal that they lack a full understanding of visual scenes.

\begin{figure*}[b]
    \centering
    \includegraphics[width=2\columnwidth]{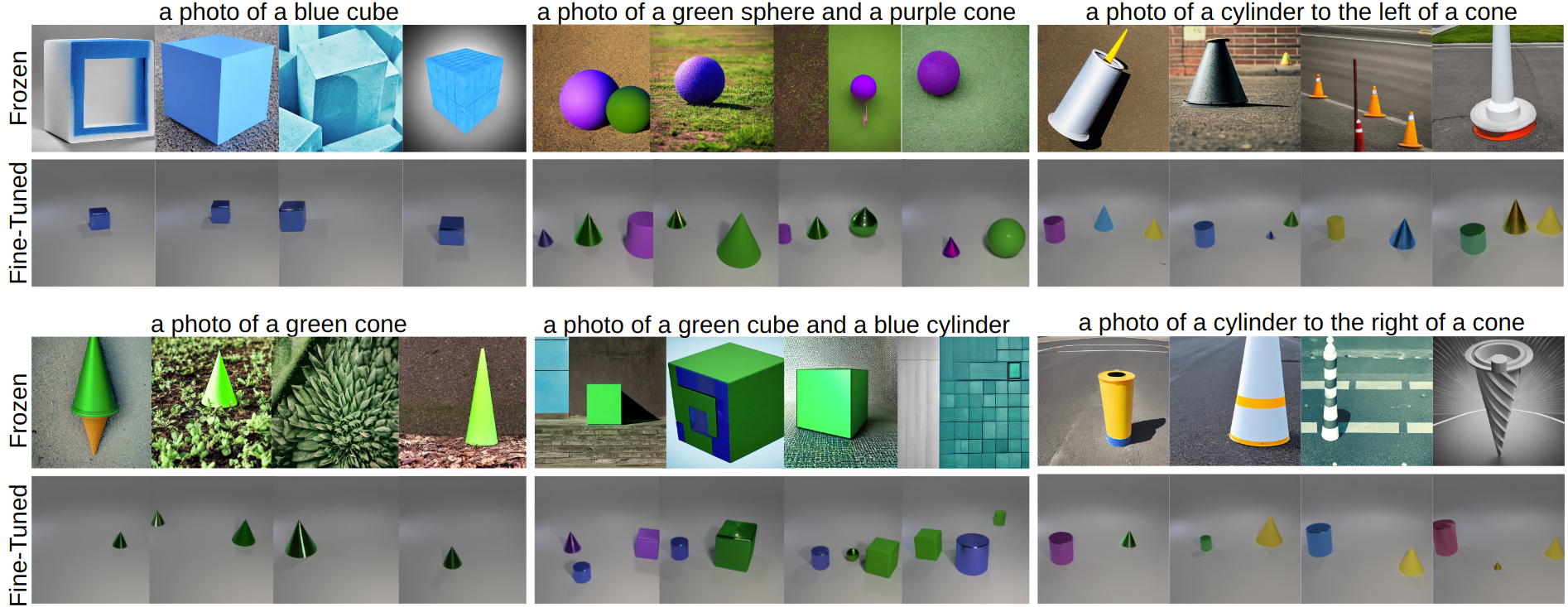}
    \caption{Images generated by Frozen and Fine-Tuned Diffusion-Classifier using prompts from the single, two-object and relational, shown from left to right. The top two rows are generated by labels from the train set and the bottom two from the test set.}
    \label{fig:diffusion-images}
\end{figure*}

\subsection{Relational Generalised Zero-Shot}
We show the performance of the models on the GZSL relational task in Table~\ref{tab:relational-generalised-zero-shot} reporting the mean and standard deviation for the fine-tuned models. We use the same fine-tuned models as in the relational ZSL setting. 

 \begin{table}[htbp]
  \begin{center}
  \resizebox{\columnwidth}{!}{%
    \begin{tabular}{lccccccr}
    \toprule
    \textbf{Model} & \textbf{ID Validation} & \textbf{ID Test} & \textbf{OOD Validation} & \textbf{OOD Test}\\\midrule
    Frozen CLIP & 27.27\textsuperscript{0.00}& 27.43\textsuperscript{0.00} & 18.00\textsuperscript{0.00} & 23.75\textsuperscript{0.00}
    \\
    CLIP-FT & 62.12\textsuperscript{0.43} & 72.22\textsuperscript{3.97}& 42.80\textsuperscript{18.39}& 34.75\textsuperscript{16.33}\\
    Frozen ViLT & 13.94\textsuperscript{0.43} & 16.55\textsuperscript{0.65} & 22.53\textsuperscript{0.19} & 26.5\textsuperscript{0.35}\\
    ViLT-FT & 16.55\textsuperscript{0.65} & 22.53\textsuperscript{0.19} & 26.5\textsuperscript{0.35} & 25.50\textsuperscript{1.08}\\
    Frozen DC & 24.55\textsuperscript{0.00} & 21.53\textsuperscript{0.00} & 10.00\textsuperscript{0.00}& 24.50\textsuperscript{0.00}\\
    DC-FT & 32.73\textsuperscript{2.57} & 34.72\textsuperscript{2.60} & 41.20\textsuperscript{4.57}& 38.25\textsuperscript{4.02}\\ 
    \bottomrule
    \end{tabular}
    }
    \caption{Accuracy of models on the GZS relational task.}
    \label{tab:relational-generalised-zero-shot}
  \end{center}
\end{table}

In the GZSL relational setting, CLIP-FT performs reasonably well on the ID splits with 62.13\% and 72.22\%, however, there is a significant drop in performance for the OOD splits with 42.80\% and 34.75\% on validation and test respectively. CLIP therefore seems to overfit to the training data and is not able to generalise to unseen labels. ViLT struggles with this task, with even the fine-tuned model hardly performing better than chance at 20\%. Interestingly, DC has a lower accuracy on the ID splits than the OOD splits. Given DC's reasonable accuracies of 89.09\% and 92.94\% in the ID ZSL experiment, it appears DC is particularly confused by the presence of hard-negative labels showing it is lacking fine-grained understanding. All models have a drop in performance from the ZSL task showing they struggle to compose relational concepts and especially can't tell the difference between hard negatives in the GZSL setting such as \emph{sphere left cube} and \emph{sphere right cube}. This suggests that the models are relying on object recognition rather than understanding relational positions which may be not have been required to be learned during pre-training.

\section{Model Understanding}
\paragraph{Stable Diffusion Images}
We compare images generated by frozen and fine-tuned Stable Diffusion to evaluate what features Diffusion Classifier is able to learn from fine-tuning on each dataset. We use a guidance scale of 7 and 50 inference steps. Examples using prompts from each dataset are shown in Figure~\ref{fig:diffusion-images}. Frozen Stable Diffusion is generally very poor at generating images in alignment with the specified prompt, except in the single-object case. Interestingly, the two-object and relational fine-tuned Stable Diffusion generate three objects fairly frequently showing some pre-training bias and knowledge is still preserved. The relational fine-tuned model fails to understand the difference between the left and right relations with the prompts ``a cylinder to the left of a  cone'' and ``a cylinder to the right of a cone'' both resulting in images of a cylinder on the left—the class seen during training.

\paragraph{CLIP embeddings}
We show t-SNE visualisations of image and text embeddings from relational dataset examples for frozen and fine-tuned CLIP. For images, we show the embeddings of 5 samples from each class and only consider classes containing \emph{left} since the corresponding classes containing \emph{right} use the same images but with different captions.

\begin{figure*}[t]
    \centering
    \begin{subfigure}[t]{0.48\textwidth}
        \centering
        \includegraphics[width=\linewidth]{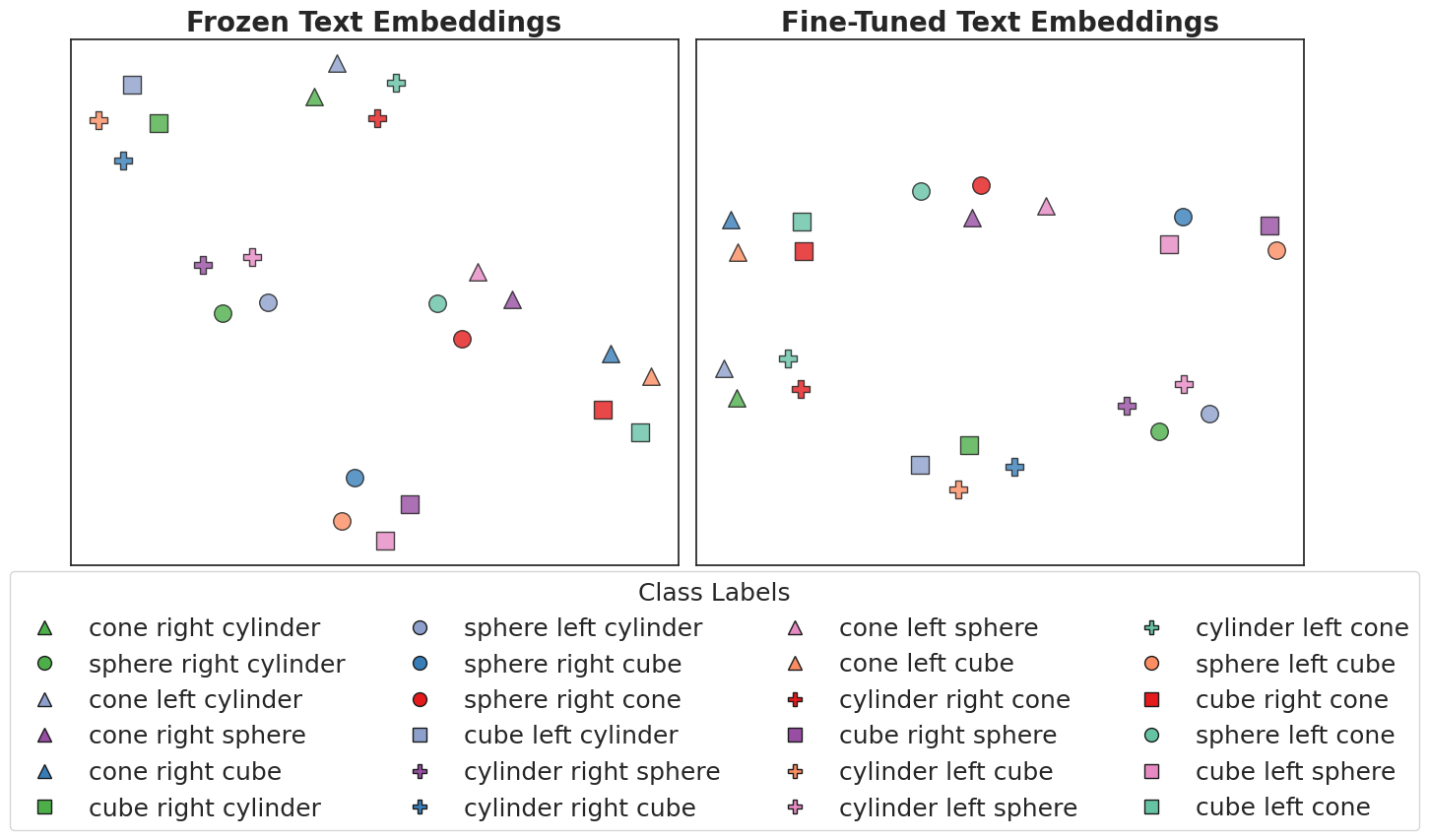}
        \caption{t-SNE visualisation of frozen and fine-tuned CLIP text embeddings for relational prompts. Best viewed electronically or in colour.}
        \label{fig:text_tsne}
    \end{subfigure}%
    \hfill
    \begin{subfigure}[t]{0.48\textwidth}
        \centering
        \includegraphics[width=\linewidth]{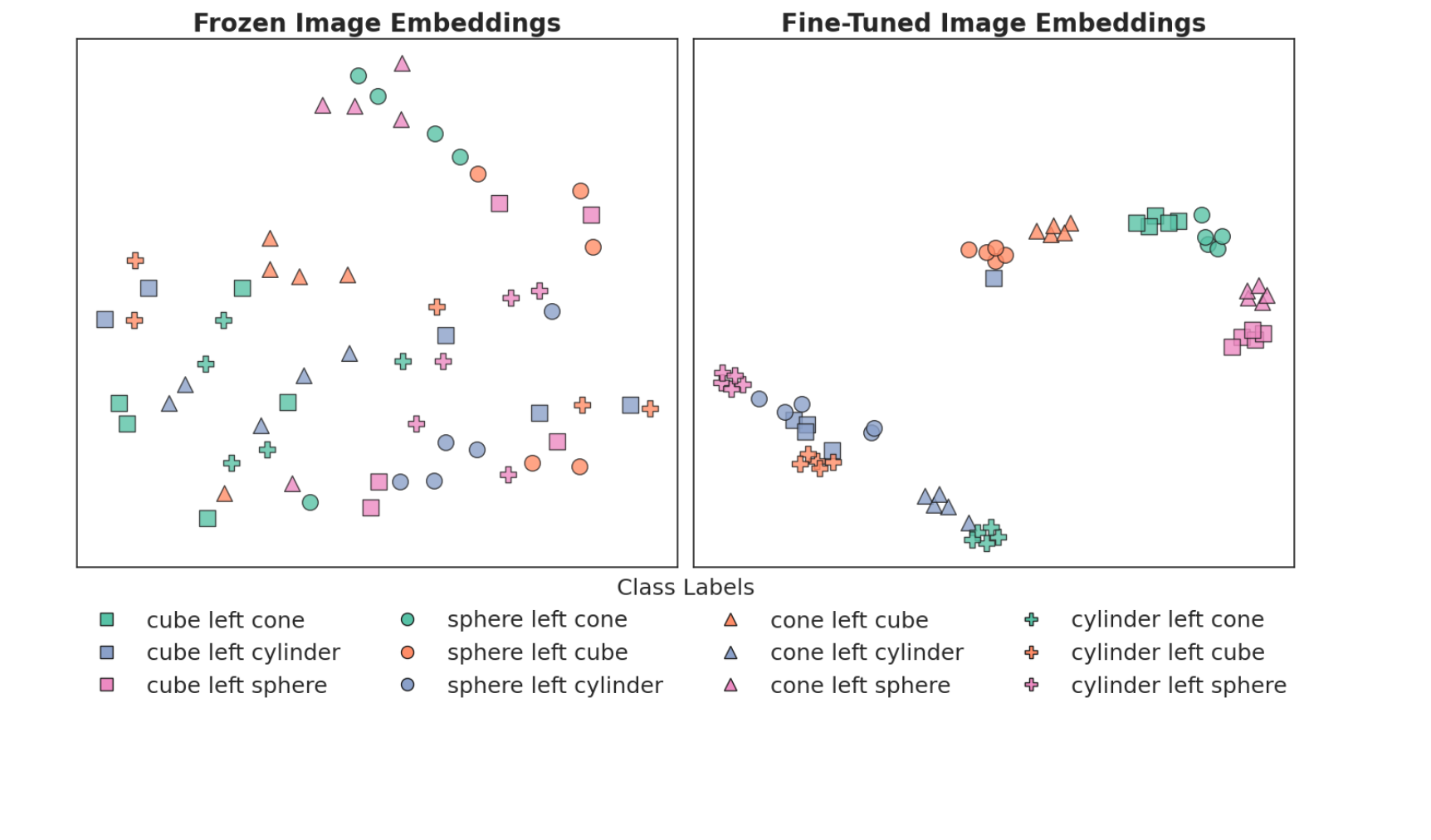}
        \caption{t-SNE visualisation of frozen and fine-tuned CLIP image embeddings for relational prompts. Best viewed electronically or in colour.}
        \label{fig:image_tsne}
    \end{subfigure}
    
    \caption{t-SNE visualisations of CLIP text and image embeddings for relational prompts after fine-tuning.}
    \label{fig:tsne_combined}
\end{figure*}

Figure~\ref{fig:text_tsne} shows the text embeddings which are clearly clustered into quadruples corresponding to prompts where the object shapes are the same, with no clear separation between prompts corresponding to different arrangements of objects. For example, the closest neighbours of \emph{cube left sphere} are \emph{sphere left cube}, \emph{cube right sphere} and \emph{sphere right cube}. Fine-tuning (right-hand plot) fails in most cases to overcome this clustering of similar prompts. An exception is the cluster of prompts containing \emph{sphere left cube} and \emph{cube right sphere}, which have been moved closer together, and are visibly distinct from \emph{sphere right cube} and \emph{cube left sphere}. Other groups of prompts tend to cluster according to ordering of nouns (e.g. \emph{cube left cone} and \emph{cube right cone}), or by bag-of-words similarity (e.g. \emph{cube left cylinder} and \emph{cylinder left cube}). This inability to distinguish prompts corresponding to different arrangements of objects likely contributes towards CLIP's inability to correctly caption images with the same shapes but different relations. 

The t-SNE visualisation of image embeddings presented in Figure~\ref{fig:image_tsne} shows that images belonging to the same class are mostly well-clustered. However, there are a few instances of classes in the wrong cluster e.g. a \textit{cube left cylinder} sample appears within the \textit{sphere left cube} cluster. Notably, we observe that embeddings of images with reversed relational structures tend to occupy similar regions in the space—for instance, \textit{cylinder left cone} and \textit{cone left cylinder} appear close together at the bottom of the plot, while \textit{cylinder left sphere} and \textit{sphere left cylinder} are both near the left side of the plot. This spatial overlap may contribute to CLIP’s difficulty with relational reasoning. Some classes, such as \textit{cube left cylinder} and \textit{sphere left cylinder}, appear slightly less well clustered. We note that these are out-of-distribution (OOD) classes, so their embeddings were not directly refined during fine-tuning.

\section{Discussion}
We extend the Concept Binding Benchmark from~\citet{lewis-etal-2024-clip} to assess concept binding in zero-shot (ZSL) and generalised zero-shot (GZSL) settings. Using this extended framework, we compare the performance of the discriminative models CLIP and ViLT against a generative model, Diffusion Classifier, on single-object, two-object, and relational compositional tasks. Diffusion Classifier shows the highest generalisation accuracy on the single-object task. ViLT achieves state-of-the-art performance on both ZSL and GZSL two-object tasks, demonstrating strong compositional ability in binding attributes to objects even in GZSL settings. Diffusion Classifier shows some capacity to generalise in the two-object GZSL setting, however, it falls short of ViLT’s performance.

On the relational composition task, all models perform poorly, showing considerable drops in performance on the GZSL from the ZSL task showing that hard distractors such as \emph{cube left sphere} versus \emph{cube right sphere} are a particular problem. Despite initial hopes that Diffusion Classifier’s generative approach might better handle compositionality, relational reasoning remains a major challenge for all models tested.

On all our experiments, our fine-tuned CLIP model consistently outperforms the model from Lewis et al. on the OOD splits~\cite{lewis-etal-2024-clip}. We attribute this to our fine-tuning strategy of only using positive examples unlike Lewis et al. who use both positive and negative examples. We hypothesise that the inclusion of negative examples leads to over-fitting. This is due to prompts appearing as negative training examples which then appear as positive examples in the OOD splits, causing CLIP to suppress their prediction. Therefore our positive-only approach appears to lead to better generalisation and reduced over-fitting.

The low performance on the GZSL relational task suggests current VLMs may rely too heavily on shortcuts such as object recognition rather than developing structured, compositional representations. Our analysis of image and text embeddings in CLIP further supports this: relational concepts (e.g., \emph{left} vs. \emph{right}) are not sufficiently disentangled, especially in the text embedding space, limiting the models’ capacity to reason about spatial relationships. Potential avenues to address this are training on datasets with more explicit compositional objectives and developing better prompting or fine-tuning strategies that encourage attribute and relation disentanglement.

However, while these routes to improved compositional understanding are important, we argue that our results highlight an important limitation of the tested models as they stand: at present compositional understanding is clearly limited. Since there may be a number of aspects of composition that we require models to perform, these should be considered at the pre-training stage rather than expecting users to fine-tune for these fundamental semantic abilities.

\section*{Limitations}
While our benchmark uses synthetic, simplistic images, we chose this design specifically to reduce the risk of spurious correlations~\cite{wu23a} and enable precise compositional structures to be tested for. We view this benchmark as a diagnostic test for probing specific compositional generalisation properties in VLMs that may be masked in more complex, real-world scenarios. Future work could include expanding these experiments to test other attributes such as material or size. Another interesting avenue for future research would be to expand the experiments to include more than two-objects. 

\section*{Acknowledgements}
The authors wish to acknowledge and thank the financial support of the UK Research and Innovation (UKRI) [Grant ref EP/S022937/1] and the University of Bristol. This work was carried out using the computational facilities of the Advanced Computing Research Centre, University of Bristol---http://www.bristol.ac.uk/acrc/

\bibliographystyle{acl_natbib}
\bibliography{custom}

\appendix
\section{Fine-tuning Details}
\label{app:fine-tune}
Optimal hyper-parameters were selected by performing a search for each model. We consider the parameters: learning rate, images per class, epochs and LoRA parameters where applicable. We select final parameters based on averaged performance on the ID val and OOD val dataset splits.

\paragraph{Single-Object} CLIP was fine-tuned using 40 images per class for 30 epochs, using an Adam optimiser with a learning rate of $1 \cdot 10^{-6}$, a batch size of 16, and a contrastive loss. For DC, we used DreamBooth to fine-tune Stable Diffusion's U-Net and text-encoder. We use 30 images per class for 4000 steps with a learning rate of $5 \cdot 10^{-6}$ and a batch size of 1. All inferences were performed using 200 noise samples. ViLT was fine-tuned on 80 images per class using LoRA with a learning rate of $1 \cdot 10^{-5}$  setting the LoRA rank (r) to 16 and the scaling factor (alpha) to 32.

\paragraph{Two-Object} CLIP was fine-tuned using 40 images per class for 30 epochs, using an Adam optimiser with a learning rate of $1 \cdot 10^{-6}$, a batch size of 16, and a contrastive loss.
For DC, we fine-tuned using 30 images per class for 4000 steps with a learning rate of $5 \cdot 10^{-6}$ and a batch size of 1. All inferences were performed using 200 noise samples. ViLT was fine-tuned using LoRA with a learning rate of $1 \cdot 10^{-5}$  setting the LoRA rank (r) to 16 and the scaling factor (alpha) to 32.

\paragraph{Relational} CLIP uses the same parameters as the single-object model except using 20 images per class for 50 epochs. DC is fine-tuned on 40 images per class for 5000 steps with the remaining parameters the same as the previous two models. ViLT is fine-tuned on 40 images per class using LoRA  with a learning rate of $1 \cdot 10^{-6}$  setting the LoRA rank (r) to 8 and the scaling factor (alpha) to 16.